\documentclass{article} 
\usepackage{iclr2027_conference,times}

\usepackage{amsmath,amsfonts,bm}

\def\eqref#1{equation~\ref{#1}}

\def\1{\bm{1}}

\DeclareMathAlphabet{\mathsfit}{\encodingdefault}{\sfdefault}{m}{sl}
\SetMathAlphabet{\mathsfit}{bold}{\encodingdefault}{\sfdefault}{bx}{n}

\usepackage{graphicx}
\usepackage{hyperref}
\usepackage{url}
\usepackage{float}
\usepackage{booktabs} \usepackage{multirow}
\title{Dual-GNN Multilevel Coarsening for Maximum Independent Set}

\author{
  Tianfeng Chen \\
  School of Mathematics and Statistics \\
  Lanzhou University\\
  \texttt{chentf2025@lzu.edu.cn}
  \And
  Xianyue Li* \\
  School of Mathematics and Statistics \\
  Lanzhou University\\
  \texttt{lixianyue@lzu.edu.cn} 
}

\iclrfinalcopy 
\begin{document}

\maketitle
\lhead{}
\begin{abstract}
The maximum independent set (MIS) problem is a fundamental NP-hard combinatorial optimization problem with applications in scheduling, resource allocation, and network analysis. Exact solvers can provide high-quality solutions or optimality certificates, but their computational cost grows rapidly with graph size, while hand-crafted heuristics improve scalability at the expense of guarantees. Learning-based methods offer an alternative by exploiting structural patterns across graph instances, yet directly predicting independent sets can make global coordination difficult on large graphs. We instead use learning to guide multilevel graph coarsening while retaining combinatorial search for final decision making. Our Dual-GNN Multilevel Coarsening framework uses a Partition GNN to score candidate contractions and a Representative GNN to select top-k local independent-set states for each final cluster. Experiments on Erd\H{o}s--R\'enyi graphs with up to 2,000 vertices demonstrate a favorable quality--runtime trade-off. On 500-vertex instances with certified optima, our method achieves an average independent-set size of 19.20, corresponding to 99.5\% of the optimal value of 19.30, while reducing the mean wall-clock time from 643.57 seconds for exact solving to 3.41 seconds, yielding an approximately 189$\times$ speedup. On larger graphs with 1,000 and 2,000 vertices, our method achieves the best mean solution quality among all evaluated methods. Moreover, although trained only on Erd\H{o}s--R\'enyi graphs with edge
probability $p=0.35$, the learned coarsening policy generalizes
effectively across both unseen graph densities and structurally
different graph families.

\end{abstract}

\section{Introduction}
\label{sec:introduction}

The MIS problem asks for the largest subset
of vertices in a graph such that no two selected vertices are adjacent.
As a fundamental NP-hard combinatorial optimization problem, MIS arises
in applications including scheduling, resource allocation, network
design, and conflict-free selection. It is also closely related to other
canonical graph problems, such as maximum clique and minimum vertex
cover. Despite its simple formulation, solving MIS becomes increasingly
difficult as the graph grows, since a locally promising vertex decision
may exclude many combinations that would lead to a better global solution.

Existing approaches to MIS can be broadly divided into exact and heuristic methods. Exact MIS algorithms typically explore a branch-and-bound search tree,
often interleaving branching with problem-specific reduction rules in
the branch-and-reduce paradigm
\citep{xiao2017exact}. Such methods are highly effective on many practical instances, but exact optimization generally requires combinatorial search whose cost grows rapidly with graph size and structural complexity.
 By contrast, specialized heuristic solvers combine kernelization, evolutionary search, and local improvement to find large independent sets on graphs where exact optimization becomes computationally expensive. These methods often achieve strong empirical performance, but their reduction, search, and refinement strategies are largely hand-designed and may require substantial computation to adapt to different graph distributions.

Learning-based combinatorial optimization offers a complementary
direction. Graph neural networks (GNNs) have been used to predict
vertex-level solution likelihoods and guide tree search
\citep{li2018combinatorial}, to optimize distributions over feasible
sets without supervised labels \citep{karalias2020erdos}, and to guide
recursive decisions through self-training
\citep{brusca2023maximum}. By learning from families of related graph
instances, these approaches can capture structural patterns that are
difficult to encode using fixed heuristics. Nevertheless, constructing a globally consistent independent set from local predictions remains challenging, as selecting one vertex can affect the feasibility and utility of many subsequent choices. Moreover, directly applying a GNN together with a search procedure to the original graph does not fundamentally reduce the computational burden associated with optimization over a large decision space. Recent evaluations of learning-guided tree search for MIS further
indicate that strong performance may depend substantially on classical
algorithmic components such as kernelization, rather than on learned
predictions alone \citep{boether2022whats}. This observation motivates
architectures in which learning and combinatorial optimization have
clearly separated and measurable roles.

To reduce the cost of combinatorial search, we propose
\emph{Dual-GNN Multilevel Coarsening}, a task-aligned framework that
contracts the original graph into a smaller optimization problem.
Unlike conventional coarsening methods that primarily preserve spectral
or structural properties
\citep{jin2020graph}, our method learns
coarsening decisions from their effects on the recovered MIS objective.
A Partition GNN scores candidate contractions between clusters, while a
Representative GNN ranks the local independent-set states retained for
coarse optimization. The resulting reduced problem
is solved using completion-aware search, after which the selected coarse
solution is decoded and completed on the original graph to produce a
valid maximal independent set.

Experiments on Erd\H{o}s--R\'enyi and structured graph families
demonstrate favorable quality--runtime trade-offs across graph sizes and
densities. On 500-vertex instances with certified optima, our method
obtains an average independent-set size of $19.20$, corresponding to
$99.5\%$ of the optimal value of $19.30$, while reducing the mean
empirical wall-clock time from $643.57$ to $3.41$ seconds. Across graph
sizes from 300 to 2,000 vertices, it consistently outperforms the
evaluated learning-based baselines in mean solution quality.

Our main contributions are:
\begin{itemize}
    \item We introduce a dual-GNN multilevel framework that learns both
    task-aligned cluster contractions and representative-state selection
    for MIS.

    \item We formulate coarsening as a bounded-state compression procedure
that preserves multiple feasible local independent-set configurations
within each cluster, enabling global compatibility to be resolved on a
substantially smaller conflict graph.

    \item We demonstrate substantial empirical runtime reductions and
    near-optimal solution quality across multiple graph sizes, densities,
    and topology families.
\end{itemize}

\section{Related Work}
\label{sec:related_work}

\paragraph{Heuristic MIS solvers.}
A simple approach to MIS is greedy construction, which repeatedly
selects a feasible vertex according to a fixed or dynamically updated
priority. Greedy algorithms are efficient and always return a maximal
independent set, but their solution quality depends strongly on the
selection rule. More advanced solvers combine reductions with global
search and local improvement. ReduMIS, for example, integrates
kernelization with evolutionary search to compute high-quality
independent sets on large graphs \citep{lamm2017finding}. Another line
of work formulates MIS as continuous optimization. pCQO-MIS optimizes a
non-convex clique-informed quadratic objective from multiple
initializations and converts the resulting solutions into maximal
independent sets \citep{alkhouri2024differentiable}. Unlike
distribution-trained models, pCQO-MIS performs differentiable
optimization separately for each input instance.

\paragraph{Learning-based MIS solvers.}
Learning-based methods use GNNs to capture recurring structural patterns
across graph instances. GCN-TreeSearch predicts vertex-level solution
likelihoods and uses them to guide tree search
\citep{li2018combinatorial}. Erd\H{o}s Goes Neural learns a distribution
over feasible sets through an unsupervised objective followed by
deterministic decoding \citep{karalias2020erdos}. Dynamic MIS instead
uses a GNN to guide recursive subproblem selection and improves the
policy through self-training \citep{brusca2023maximum}. Our approach
differs from these methods by applying learning to multilevel
coarsening: the two GNNs determine which clusters to contract and which
local states to retain before the reduced combinatorial search.

\paragraph{Graph coarsening.}
Graph coarsening became widely established through multilevel graph
partitioning, where a graph is repeatedly contracted, partitioned at
the coarsest level, and then uncoarsened and refined
\citep{karypis1998fast}. More recent methods primarily aim to preserve
spectral or structural properties
\citep{loukas2018spectrally,jin2020graph}, while Cai et al.\ use a GNN
to learn coarse edge weights under general preservation objectives
\citep{cai2021graph}. Such objectives are not necessarily aligned with
MIS, since a structurally reasonable contraction may remove useful
independent-set configurations. Our method instead learns contraction
and representative-state scores from rollout-based estimates of the
final MIS quality.

\section{Preliminaries}
\label{sec:preliminaries}

\paragraph{Maximum independent set.}
Let $G=(V,E)$ be an undirected graph with vertex set $V$ and edge set
$E$. A subset $I\subseteq V$ is an independent set if no two vertices
in $I$ are adjacent. The maximum independent set problem seeks an
independent set of maximum cardinality. Using a binary variable
$x_v\in\{0,1\}$ to indicate whether vertex $v$ is selected, the problem
can be formulated as
\begin{equation}
\begin{aligned}
\alpha(G)
&= \max_{\mathbf{x}\in\{0,1\}^{|V|}}
\sum_{v\in V} x_v \\
\text{s.t.}\quad
& x_u+x_v\leq 1,
\qquad \forall (u,v)\in E .
\end{aligned}
\label{eq:mis}
\end{equation}
An independent set is \emph{maximal} if no additional vertex can be added
without violating independence.

\paragraph{Graph neural networks.}
A graph neural network constructs a representation
$\mathbf{h}_v^{(k)}$ for each vertex by repeatedly aggregating
information from its neighbors \citep{kipf2016semi,hamilton2017inductive}.
A generic message-passing layer can be written as
\begin{equation}
    \mathbf{m}_v^{(k)}
    =
    \operatorname{AGG}^{(k)}
    \left(
        \left\{
            \mathbf{h}_u^{(k-1)}
            : u\in\mathcal{N}(v)
        \right\}
    \right),
    \qquad
    \mathbf{h}_v^{(k)}
    =
    \operatorname{UPDATE}^{(k)}
    \left(
        \mathbf{h}_v^{(k-1)},
        \mathbf{m}_v^{(k)}
    \right),
    \label{eq:message_passing}
\end{equation}
where $\mathcal{N}(v)$ denotes the neighborhood of $v$. In this work,
GNNs are applied not only to the original graph but also to quotient
graphs, where each node represents a cluster of original vertices.

\paragraph{Multilevel graph coarsening.}
Graph coarsening constructs a sequence of progressively smaller graphs
by grouping vertices into clusters. At level $\ell$, let
\begin{equation}
    \mathcal{C}^{(\ell)}
    =
    \left\{
        C_1^{(\ell)},\ldots,C_{n_\ell}^{(\ell)}
    \right\}
\end{equation}
be a partition of the original vertex set. Each cluster is represented
as a node in the quotient graph $G^{(\ell)}$, and two coarse nodes are
connected when their corresponding clusters share at least one edge in
the original graph. A contraction combines two disjoint clusters,
\begin{equation}
    C_{ij}^{(\ell+1)}
    =
    C_i^{(\ell)}\cup C_j^{(\ell)}.
\end{equation}
Contracting a set of non-overlapping pairs produces the next level of
the hierarchy,
\begin{equation}
    G^{(0)}
    \rightarrow G^{(1)}
    \rightarrow \cdots
    \rightarrow G^{(L)},
    \qquad G^{(0)}=G,
    \label{eq:coarsening_hierarchy}
\end{equation}
with $|V^{(\ell+1)}|<|V^{(\ell)}|$. Classical multilevel methods select
contractions using structural heuristics
\citep{karypis1998fast}. Our method instead uses learned, task-aligned
scores to construct the hierarchy and preserve local decisions that are
useful for the final MIS solution.

\section{Dual-GNN Multilevel Coarsening}
\label{sec:method}

\subsection{Framework Overview}
\label{sec:overview}
\begin{figure}
    \centering
    \includegraphics[width=1.0\linewidth]{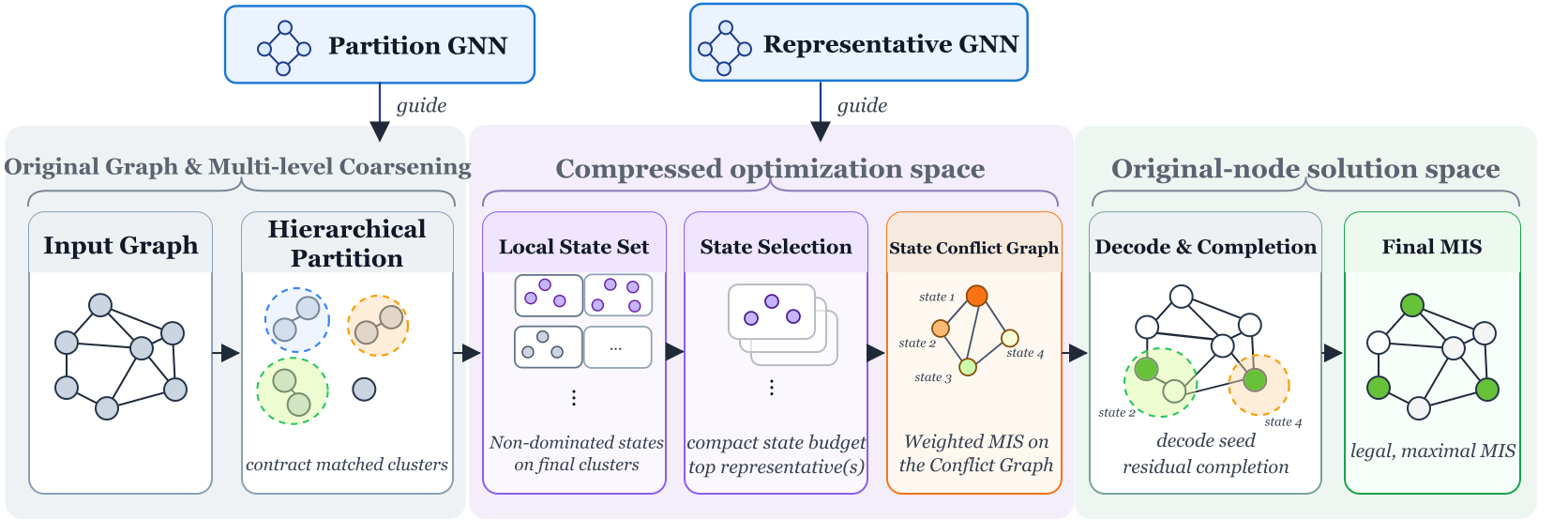}
    
    \caption{Overview of the proposed Dual-GNN Multilevel Coarsening framework}
    \label{fig:framework}
\end{figure}

Figure~\ref{fig:framework} provides an overview of our framework. Given
an input graph $G=(V,E)$, the method first constructs a multilevel
coarsening hierarchy. Initially, each vertex forms a singleton cluster.
At every level, the current clusters are represented as nodes in a
quotient graph, whose edges summarize the connectivity between clusters.
The Partition GNN encodes this quotient graph and scores candidate
cluster pairs. Based on these scores, a set of non-overlapping pairs is
selected as a matching and contracted to form the next coarsening level.
This process is repeated until a stopping criterion is reached, while a
maximum cluster size limits the complexity of subsequent local
optimization.

After constructing the hierarchy, we enumerate feasible local
independent-set states within each final cluster. The Representative GNN
evaluates these states using both their local properties and the context
provided by the final quotient graph. Only a bounded set of
representative states is retained for each cluster. The retained states
define a coarse conflict graph: each node represents a local independent
set, its weight is the number of selected original vertices, and an edge
indicates that two states are mutually exclusive or contain conflicting
vertices in the original graph. Consequently, the size of the coarse
search space is controlled explicitly by the representative-state
budget rather than by the total number of locally enumerated states.

The coarse conflict graph is then processed by a bounded combinatorial
search to obtain candidate seeds. Since the size of a coarse seed does
not fully reflect its potential on the original graph, we use
completion-aware evaluation to rank candidates according to the
independent sets obtained after decoding and residual completion. The
best candidate is mapped back to its original vertices and
deterministically completed to a maximal independent set. 

\subsection{Partition GNN}
\label{sec:partition_gnn}

\begin{figure}
    \centering
    \includegraphics[width=1.0\linewidth]{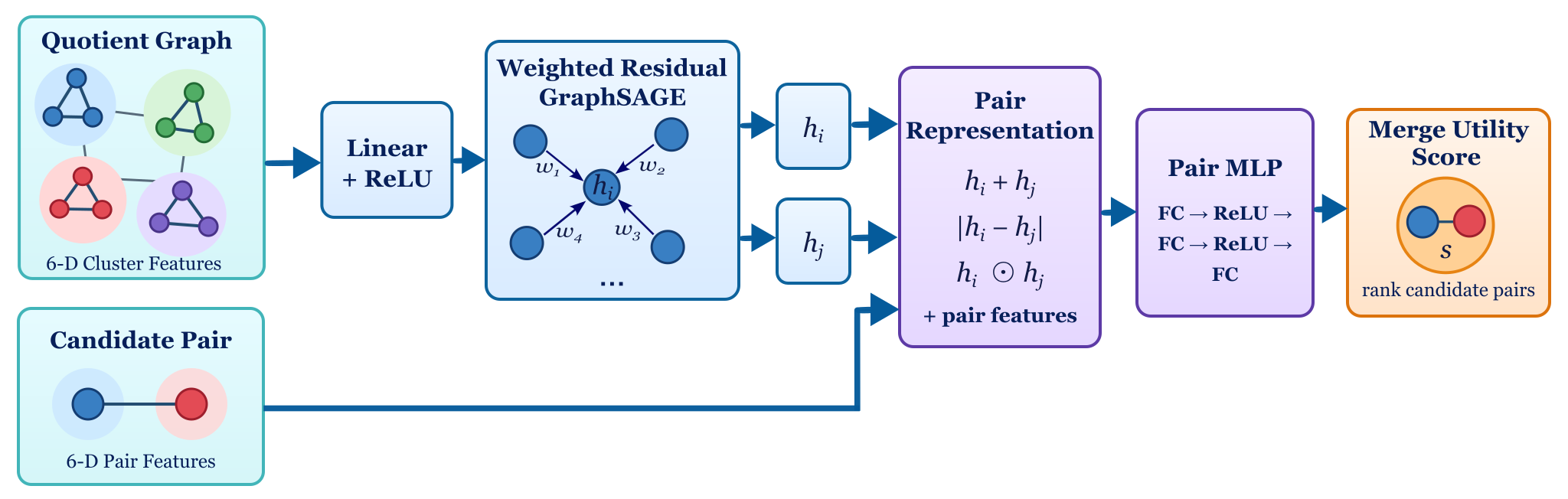}
    \caption{Partition GNN–Guided Cluster Contraction Pipeline}
    \label{fig:gnn1}
\end{figure}

The Partition GNN determines which clusters should be contracted at
each level of the coarsening hierarchy. Let
$\mathcal{C}^{(\ell)}=\{C_1,\ldots,C_{m_\ell}\}$ denote the current
partition of the original vertex set, and let
$Q^{(\ell)}$ be the corresponding quotient graph. Our
partitioning stage maintains only the vertex membership of each
cluster. Local independent-set states are generated after the final
coarsening level, which avoids an exponential growth of intermediate
states. Figure~\ref{fig:gnn1} provides an overview of the Partition GNN pipeline for scoring candidate cluster contractions. 

Each quotient node is represented by a six-dimensional feature vector
\begin{equation}
\mathbf{x}_i =
\left[
\frac{|C_i|}{n},\;
\frac{|\partial C_i|}{n},\;
\frac{d_Q(i)}{m_\ell-1},\;
\rho(C_i),\;
\frac{\bar d(C_i)}{n},\;
\frac{d_{\min}(C_i)}{n}
\right].
\label{eq:partition_node_features}
\end{equation}
Here, $\partial C_i$ is the external boundary of $C_i$,
$d_Q(i)$ is its degree in the quotient graph, $\rho(C_i)$ is its
internal edge density, and $\bar d(C_i)$ and $d_{\min}(C_i)$ are the
mean and minimum original degrees of the vertices in the cluster. To distinguish weak and strong interactions between
clusters, an edge between $C_i$ and $C_j$ is assigned the normalized
weight
\begin{equation}
\omega_{ij}
=
\frac{|E(C_i,C_j)|}{|C_i||C_j|},
\label{eq:quotient_edge_weight}
\end{equation}
where $E(C_i,C_j)$ denotes the set of original edges crossing between
the two clusters.

We apply residual GraphSAGE layers to the weighted quotient graph to
obtain a contextual embedding $\mathbf{h}_i$ for every cluster.
For each admissible pair $(C_i,C_j)$, we additionally construct a
pairwise feature vector $\mathbf{p}_{ij}$ containing the cross-edge
density, boundary Jaccard similarity, cluster-size balance, quotient
degree difference, merged boundary size and merged cluster size. The learned contraction utility is
computed as
\begin{equation}
u_{ij}
=
\operatorname{MLP}_{P}
\left(
\left[
\mathbf{h}_i+\mathbf{h}_j
\;\middle\|\;
|\mathbf{h}_i-\mathbf{h}_j|
\;\middle\|\;
\mathbf{h}_i\odot\mathbf{h}_j
\;\middle\|\;
\mathbf{p}_{ij}
\right]
\right).
\label{eq:partition_pair_score}
\end{equation}
The sum, absolute difference, and element-wise product make
$u_{ij}$ invariant to the ordering of the two clusters.

Only pairs satisfying $|C_i|+|C_j|\leq B$ are considered. We define
the structural contraction utility as
\begin{equation}
u_{ij}^{\mathrm{struct}}
=
-\frac{|(\partial C_i\cup\partial C_j)\setminus(C_i\cup C_j)|}{n}
+0.35
\frac{|\partial C_i\cap\partial C_j|}
{\max(|\partial C_i\cup\partial C_j|,1)}
-0.10\omega_{ij}.
\label{eq:structural_contraction}
\end{equation}
The learned and structural utilities are combined as
\begin{equation}
q_{ij}
=
u_{ij}^{\mathrm{struct}}
+\lambda_P u_{ij},
\qquad
\lambda_P=0.5,
\label{eq:combined_contraction}
\end{equation}
where a larger value indicates a more favorable contraction. The
coefficient $\lambda_P$ is selected on the validation set and fixed for
all experiments. Candidate pairs are sorted in descending order of
$q_{ij}$, after which a greedy matching selects non-overlapping
contractions. In the variant without the Partition GNN, the learned
term is removed, and the same candidates are ranked using
$u_{ij}^{\mathrm{struct}}$ alone.

\subsection{Representative GNN}
\label{sec:representative_gnn}

\begin{figure}
    \centering
    \includegraphics[width=1.0\linewidth]{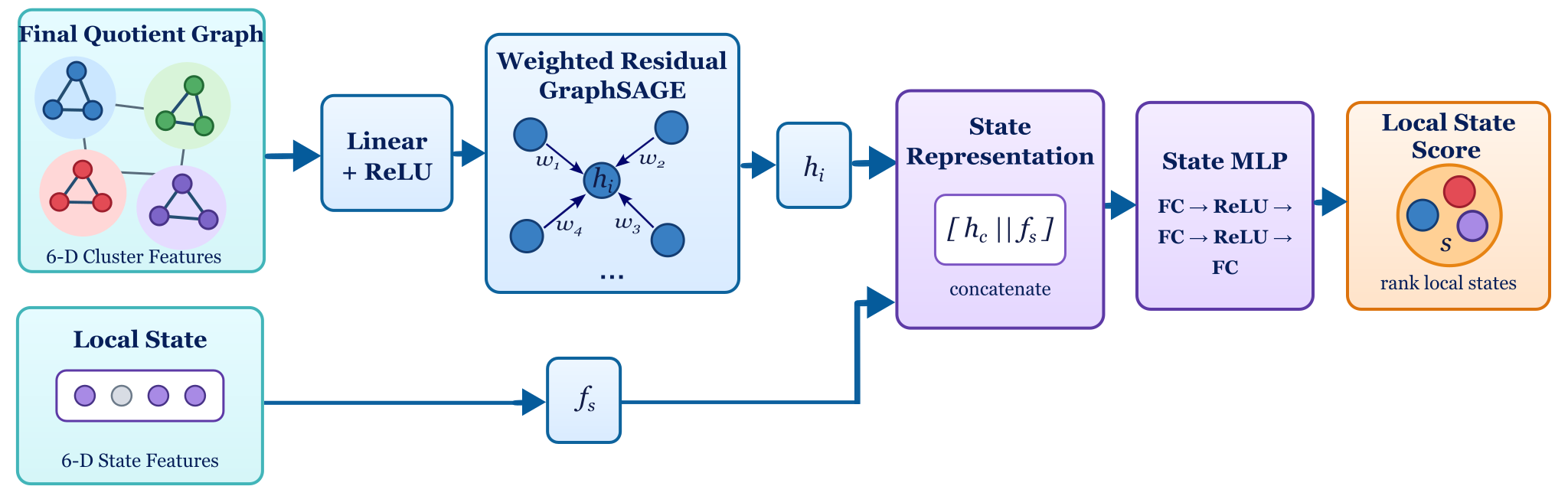}
    \caption{Representative GNN state-selection pipeline}
    \label{fig:gnn2}
\end{figure}

After the final coarsening level, each cluster may have several feasible local independent-set states. Keeping all of them would reduce the efficiency gained from coarsening. However, selecting states only by their size may remove some states that work better with the rest of the graph.
We therefore use a Representative
GNN to rank the feasible local states of each final cluster according to
both their local quality and their compatibility with the surrounding
graph. Figure~\ref{fig:gnn2} provides an overview of the Representative GNN pipeline for scoring and selecting representative local states for each final cluster.

For a final cluster $C_i$, we enumerate its non-empty local independent
sets,
\begin{equation}
    \mathcal{S}_i
    =
    \left\{
        S\subseteq C_i:
        S\neq\emptyset,\;
        (u,v)\notin E
        \ \forall u,v\in S
    \right\}.
\end{equation}
The empty state is represented implicitly by not selecting any state
from the cluster during coarse optimization. Each state
$S\in\mathcal{S}_i$ is associated with its cardinality
$w(S)=|S|$ and its external blocking set
\begin{equation}
    B(S)
    =
    \left\{
        v\in V\setminus C_i:
        \exists u\in S \text{ such that } (u,v)\in E
    \right\}.
    \label{eq:state_blocking_set}
\end{equation}
A state $S$ dominates another state $T$ if
$w(S)\geq w(T)$ and $B(S)\subseteq B(T)$, since then $S$ is not worse
for any assignment outside the cluster. States with identical blocking
sets are deduplicated, and dominated states are removed before learning
ranking.

The Representative GNN encodes the final quotient graph using the same
six-dimensional cluster descriptors introduced in
Equation~\ref{eq:partition_node_features}. Quotient edges are weighted
by their normalized cross-edge densities from
Equation~\ref{eq:quotient_edge_weight}. Residual GraphSAGE layers then
produce a context-dependent embedding $\mathbf{h}_i$ for every final
cluster. This embedding allows the state ranker to account for how the
cluster is situated relative to the other clusters rather than
evaluating each local state in isolation.

For each candidate state $S\in\mathcal{S}_i$, we construct the
six-dimensional state feature vector
\begin{equation}
\mathbf{z}_{iS}
=
\left[
\frac{|S|}{|C_i|},\;
\frac{|B(S)|}{n},\;
\frac{\bar d(S)}{n},\;
\frac{d_{\min}(S)}{n},\;
\frac{|B(S)|}{\max(|\partial C_i|,1)},\;
\frac{|S|}{w_i^{\max}}
\right],
\label{eq:representative_state_features}
\end{equation}
where $B(S)$ is the set of external vertices blocked by $S$,
$\bar d(S)$ and $d_{\min}(S)$ are respectively the mean and minimum
original-graph degrees of the vertices selected by $S$, and
\begin{equation}
    w_i^{\max}
    =
    \max_{T\in\mathcal{S}_i}|T|
\end{equation}
is the cardinality of the largest local independent-set state in
cluster $C_i$. These features describe the local solution size, its
external conflicts, the degree characteristics of its selected
vertices, and its quality relative to the other states of the same
cluster. The Representative GNN combines these local features with the
embedding of the corresponding cluster:
\begin{equation}
    r_{iS}
    =
    \operatorname{MLP}_{R}
    \left(
        \left[
            \mathbf{h}_i
            \;\middle\|\;
            \mathbf{z}_{iS}
        \right]
    \right),
    \label{eq:representative_state_score}
\end{equation}
where $\mathbf{z}_{iS}$ denotes the state-level feature vector and
$r_{iS}$ is the learned representative score.

States are ranked independently within each cluster according to
$r_{iS}$. We retain the top-$k$ states in each cluster, where $k=3$ is
selected on the validation set. A sensitivity analysis of $k$ is
provided in Appendix~\ref{fig:placeholder}. The retained states
form the nodes of the subsequent coarse conflict graph.

\subsection{Coarse Optimization and Solution Recovery}
\label{sec:coarse_optimization}

The retained states define a weighted coarse conflict graph. Each node
represents a local state $S$ and has weight $|S|$. Two states are
connected if they belong to the same cluster or if their selected
vertices conflict in the original graph.
We then solve a weighted MIS problem on this conflict graph
to obtain a feasible coarse seed.

Since the coarse objective does not account for vertices that can be
added after decoding, candidate seeds are evaluated using
\begin{equation}
F(I_{\mathrm{seed}})
=
\left|
\operatorname{Complete}(I_{\mathrm{seed}},G)
\right|.
\end{equation}

Finally, the best seed is mapped back to its original vertices and
greedily completed on the residual graph. The resulting solution is
independent by construction and maximal because completion terminates
only when no additional feasible vertex can be added.

\subsection{Training the Dual GNNs}
\label{sec:training}

Both GNNs are trained using rollout-derived supervision. For the
Partition GNN, we generate several candidate matchings at each
coarsening level, execute the remaining coarsening and solution-recovery
pipeline, and use the final MIS size as the utility of each matching.
The network is trained with a ranking loss that assigns higher scores
to matchings producing larger recovered independent sets.

For the Representative GNN, each feasible local state is fixed and
greedily completed on the residual graph. The size of the completed
independent set is used as its rollout utility. These utilities define
soft ranking targets, and the network is trained using cross-entropy.
We train the two models alternately: the Representative GNN is trained
on hierarchies generated by the Partition GNN, after which the Partition
GNN is updated using rollouts produced with the learned representative
policy. Both models are optimized using AdamW.

\section{Experiments}
\label{sec:experiments}

\subsection{Experimental Setup}
\label{sec:experimental_setup}

\paragraph{Datasets and splits.}
We conduct the main experiments on Erd\H{o}s--R\'enyi (ER) graphs with
edge probability $p=0.35$. The training pool contains 800 graphs,
consisting of 400 graphs with 100 vertices and 400 graphs with 300
vertices. An additional set of 100 graphs with 500 vertices is reserved
for validation and model selection. For evaluation, we use 50 ER graphs for each $n \in \{100,300,500,1000,2000\}$, together with BA, SBM, and WS graphs with $n=1000$ for structural-shift evaluation.

\paragraph{Baselines.}
We compare our method with both classical and learning-based MIS
approaches. The classical baselines include a residual minimum-degree
greedy heuristic, an exact maximum-clique solver applied to the complement graph
and ReduMIS \citep{lamm2017finding}.

The learning-based baselines include Erd\H{o}s Goes Neural
\citep{karalias2020erdos}, Dynamic MIS \citep{brusca2023maximum}, and
pCQO-MIS \citep{alkhouri2024differentiable}.

\paragraph{Implementation details.}
Both the Partition GNN and the Representative GNN use two residual
GraphSAGE layers with a hidden dimension of 64.  At each coarsening
level, the number of clusters is reduced by approximately 50\%.
Coarsening terminates when the number of clusters reaches 25, when no
sufficient set of admissible contractions remains, or after at most
four coarsening rounds. The maximum cluster size is set to $B=8$ to
keep local-state enumeration tractable.

The two GNNs are trained using rollout-derived supervision on the
training split, and checkpoints are selected according to the recovered
MIS objective on the validation split. Experiments are conducted on a workstation equipped with an Intel Core
i7-10870H CPU and an NVIDIA GeForce RTX 3060 Laptop GPU with 6 GB of
memory. The implementation uses Python 3.13.5, PyTorch 2.7.1 with CUDA
11.8, PyTorch Geometric 2.6.1, NumPy 2.3.3, NetworkX 2.8.8, and
PySCIPOpt 5.6.0.

\subsection{Main Results}
\label{sec:main_results}

\paragraph{Solution Quality across Graph Sizes.}We first evaluate the solution quality of all methods on the
Erd\H{o}s--R\'enyi test sets. Each test set contains 50 independently
generated graphs with edge probability $p=0.35$. We report the mean
cardinality of the returned independent sets, where a larger value is
better. Certified optimal values are available for graphs with up to
500 vertices.

\begin{table}[H]
\centering
\caption{Solution quality on ER graphs.}
\label{tab:er_main_quality}
\setlength{\tabcolsep}{4.2pt}
\renewcommand{\arraystretch}{1.12}
\resizebox{\linewidth}{!}{%
\begin{tabular}{rccccccc}
\hline
$n$
& Optimal
& Greedy
& Erd\H{o}s G.N.
& Dynamic MIS
& pCQO-MIS
& ReduMIS
& Ours \\
\hline
100
& $12.92 \pm 0.60$
& $11.32 \pm 0.89$
& $11.72 \pm 0.76$
& $10.02 \pm 0.87$
& $\mathbf{12.92 \pm 0.60}$
& $\mathbf{12.92 \pm 0.60}$
& $\mathbf{12.92 \pm 0.60}$ \\

300
& $17.26 \pm 0.44$
& $14.94 \pm 0.82$
& $13.86 \pm 1.11$
& $12.84 \pm 1.06$
& $17.20 \pm 0.40$
& $\mathbf{17.22 \pm 0.42}$
& $\mathbf{17.22 \pm 0.42}$ \\

500
& $19.30 \pm 0.46$
& $16.20 \pm 0.73$
& $13.22 \pm 0.93$
& $14.32 \pm 0.94$
& $18.94 \pm 0.42$
& $19.12 \pm 0.33$
& $\mathbf{19.20 \pm 0.40}$ \\

1000
& --
& $18.30 \pm 0.81$
& $16.76 \pm 0.87$
& $15.48 \pm 1.01$
& $20.46 \pm 0.61$
& $21.14 \pm 0.35$
& $\mathbf{21.28 \pm 0.50}$ \\

2000
& --
& $20.32 \pm 0.71$
& $17.38 \pm 0.97$
& $16.80 \pm 1.16$
& $20.94 \pm 0.24$
& $23.14 \pm 0.35$
& $\mathbf{23.32 \pm 0.47}$ \\
\hline
\end{tabular}%
}
\end{table}

Table~\ref{tab:er_main_quality} compares the solution quality of different methods on ER graphs of varying sizes. Each entry reports the mean MIS size and sample standard deviation over the same 50 test instances. Overall, our method achieves the best or tied-best solution quality among all non-exact methods across all graph sizes. For small instances with $n=100$, our method reaches the optimal solution, matching pCQO-MIS and ReduMIS. At $n=300$ and $n=500$, it remains very close to the optimum and outperforms the other learning-based and heuristic baselines. As the graph size increases, the advantage becomes more evident. For $n=1000$ and $n=2000$, where optimal solutions are unavailable, our method obtains the largest independent sets among all compared methods, achieving average solution sizes of 21.28 and 23.32, both exceeding those obtained by ReduMIS. These results indicate that the proposed method maintains strong solution quality as the problem size grows and scales effectively to larger ER graphs.

\paragraph{Runtime Comparison.} We next compare the empirical solving time of the evaluated methods.
Table~\ref{tab:er_runtime} reports the mean wall-clock time per graph.
Offline training time is excluded. For the exact baseline, all instances with 100, 300, and 500 vertices are solved using an exact maximum-clique solver on the complement graph.
 Exact results are not available for the two
largest graph sizes.

\begin{table}[H]
    \centering
    \caption{Mean solving time on ER graphs.}
    \label{tab:er_runtime}
    \setlength{\tabcolsep}{4.2pt}
    \renewcommand{\arraystretch}{1.12}
    \resizebox{\linewidth}{!}{%
    \begin{tabular}{rccccccc}
        \hline
        $n$
        & Exact
        & Greedy
        & Erd\H{o}s G.N.
        & Dynamic MIS
        & pCQO-MIS
        & ReduMIS
        & Ours \\
        \hline
        100
        & $0.01\pm0.00$
        & $0.00\pm0.00$
        & $0.08\pm0.04$
        & $0.24\pm0.06$
        & $3.58\pm0.59$
        & $0.70\pm0.08$
        & $0.29\pm0.07$ \\

        300
        & $8.62\pm1.66$
        & $0.02\pm0.00$
        & $0.19\pm0.04$
        & $0.50\pm0.18$
        & $3.74\pm0.59$
        & $3.71\pm0.29$
        & $1.46\pm0.42$ \\

        500
        & $643.57\pm112.94$
        & $0.07\pm0.01$
        & $0.47\pm0.07$
        & $1.42\pm0.61$
        & $3.72\pm0.58$
        & $12.91\pm1.23$
        & $3.41\pm0.78$ \\

        1000
        & --
        & $0.41\pm0.04$
        & $2.14\pm0.05$
        & $10.66\pm7.50$
        & $4.16\pm0.67$
        & $22.65\pm2.33$
        & $13.12\pm3.35$ \\

        2000
        & --
        & $2.34\pm0.39$
        & $23.50\pm15.33$
        & $27.01\pm16.18$
        & $6.51\pm1.60$
        & $75.52\pm22.16$
        & $34.55\pm2.16$ \\
        \hline
    \end{tabular}%
    }
\end{table}

The exact solver becomes particularly expensive at $n=500$, requiring
643.57 seconds per graph on average. In comparison, our method requires only 3.41 seconds, yielding an approximately $189\times$ speedup while retaining 99.5\% of the optimal solution value.

For ReduMIS, the internal search time is constrained to match the solving time of our method. This limit controls only the internal search phase of
ReduMIS and does not include graph conversion, preprocessing,
initialization, process invocation, or solution parsing. We therefore
report the measured end-to-end wall-clock time for all methods. As a
result, the observed runtime of ReduMIS can exceed its nominal internal
search budget.

Greedy and the direct prediction baselines are faster in absolute
runtime, but Table~\ref{tab:er_main_quality} shows that this is
accompanied by a considerable reduction in solution quality. Similarly,
pCQO-MIS is faster than our method on the two largest graph sizes, but
returns independent sets that are smaller by 0.82 vertices at $n=1000$
and 2.32 vertices at $n=2000$. Overall, our method provides a favorable
quality--runtime trade-off: it exceeds the solution
quality of ReduMIS while requiring substantially less computation.

\paragraph{Generalization across Graph Densities.} To evaluate robustness to changes in graph density, we test the models
on ER graphs with $n=500$ and edge probabilities ranging from 0.05 to
0.70. Each test set contains 50 independently generated graphs. Our
GNNs are trained only on ER graphs with $p=0.35$ and are applied to all
other densities without retraining.

\begin{table}[H]
    \centering
    \caption{Solution quality across ER graph densities.}
    \label{tab:density-quality}
    \setlength{\tabcolsep}{4.2pt}
    \renewcommand{\arraystretch}{1.12}

    \resizebox{\linewidth}{!}{%
    \begin{tabular}{rcccccc}
        \hline
        $p$
        & Greedy
        & Erd\H{o}s G.N.
        & Dynamic MIS
        & pCQO-MIS
        & ReduMIS
        & Ours \\
        \hline

        0.05
        & $82.68 \pm 2.30$
        & $71.92 \pm 1.93$
        & $72.40 \pm 2.23$
        & \textit{Fail}
        & $\mathbf{91.56 \pm 1.36}$
        & $90.20 \pm 1.54$ \\

        0.10
        & $49.64 \pm 1.38$
        & $44.72 \pm 1.68$
        & $42.26 \pm 2.40$
        & \textit{Fail}
        & $\mathbf{55.62 \pm 0.60}$
        & $55.04 \pm 0.88$ \\

        0.20
        & $28.10 \pm 1.16$
        & $25.16 \pm 1.20$
        & $22.24 \pm 1.38$
        & $29.68 \pm 2.69$
        & $31.40 \pm 0.61$
        & $\mathbf{31.58 \pm 0.54}$ \\

        0.35
        & $16.20 \pm 0.73$
        & $13.22 \pm 0.93$
        & $14.32 \pm 0.94$
        & $18.94 \pm 0.42$
        & $19.12 \pm 0.33$
        & $\mathbf{19.20 \pm 0.40}$ \\

        0.50
        & $11.26 \pm 0.63$
        & $10.24 \pm 0.62$
        & $10.42 \pm 0.88$
        & $13.10 \pm 0.30$
        & $13.08 \pm 0.27$
        & $\mathbf{13.12 \pm 0.33}$ \\

        0.70
        & $7.14 \pm 0.53$
        & $6.94 \pm 0.42$
        & $6.76 \pm 0.52$
        & $8.46 \pm 0.50$
        & $\mathbf{8.48 \pm 0.50}$
        & $8.42 \pm 0.50$ \\

        \hline
    \end{tabular}%
    }

    \vspace{2pt}

    \begin{minipage}{\linewidth}
        \footnotesize
        \textit{Note:} At $p=0.05$ and $p=0.10$, pCQO-MIS completed all
        50 optimization batches but produced no binary assignment satisfying
        its released fixed-point maximality check under the fixed GNM
        hyperparameters; we therefore report these runs as \textit{Fail}.
    \end{minipage}

\end{table}

Table~\ref{tab:density-quality} shows that our method generalizes
effectively across a broad range of graph densities despite being
trained only at $p=0.35$. It consistently outperforms the greedy
heuristic, Erd\H{o}s Goes Neural, and Dynamic MIS. Compared with
pCQO-MIS, our method obtains better results from $p=0.20$ to $p=0.50$
and remains within 0.04 vertices at $p=0.70$.

Our method is also competitive with ReduMIS throughout the density
range. It outperforms ReduMIS at $p=0.20$, $p=0.35$ and $p=0.50$, trails it by only 0.06 vertices at
$p=0.70$. The larger gaps of 1.36 and 0.58 vertices at $p=0.05$ and
$p=0.10$ indicate that very sparse graphs remain the most challenging
density shift for the learned coarsening policy. Nevertheless, even in
this regime, our method improves substantially over the greedy and
learning-based prediction baselines.  The corresponding solving times across different graph densities are
reported in Appendix Table~\ref{tab:er_density_runtime}.

\paragraph{Generalization across Graph Structures.}
To further evaluate robustness to structural distribution shifts, we test
our method on three families of high-density synthetic graphs with
$n=1000$: Barab\'asi--Albert (BA), Stochastic Block Model (SBM), and
Watts--Strogatz (WS) graphs. These graph families respectively exhibit
heterogeneous degree distributions, community structure, and small-world
characteristics.

\begin{table}[H]
\centering
\caption{Comparison across graphs with different structural properties
at $n=1000$.}
\label{tab:graph-family-generalization}
\setlength{\tabcolsep}{3.5pt}
\renewcommand{\arraystretch}{1.12}
\resizebox{\linewidth}{!}{%
\begin{tabular}{lcccccc}
\hline
Graph
& Greedy
& Erd\H{o}s GNN
& Dynamic MIS
& pCQO-MIS
& ReduMIS
& Ours \\
\hline
BA
& $33.84 \pm 1.22$
& $29.02 \pm 1.77$
& $24.78 \pm 2.16$
& $37.29 \pm 0.98$
& $39.16 \pm 0.66$
& $\mathbf{39.64 \pm 0.68}$ \\

SBM
& $17.84 \pm 0.77$
& $16.24 \pm 0.77$
& $14.86 \pm 1.21$
& $18.95 \pm 1.34$
& $20.66 \pm 0.52$
& $\mathbf{21.04 \pm 0.20}$ \\

WS
& $8.94 \pm 0.55$
& $8.82 \pm 0.60$
& $8.78 \pm 0.72$
& $10.32 \pm 0.54$
& $11.32 \pm 0.42$
& $\mathbf{11.78 \pm 0.47}$ \\
\hline
\end{tabular}%
}
\end{table}

As shown in Table~\ref{tab:graph-family-generalization}, our method achieves the highest average MIS size on all three graph families, consistently outperforming the greedy heuristic, the direct learning-based methods, pCQO-MIS, and ReduMIS. Compared with the strongest baseline, ReduMIS, our method improves the average solution size by 0.48, 0.38, and 0.46 vertices on BA, SBM, and WS graphs, respectively. The gains over pCQO-MIS are more pronounced, reaching 2.35, 2.09, and 1.46 vertices, while the substantially larger margins over Erd\H{o}s Goes Neural and Dynamic MIS highlight the advantage of learning to coarsen the optimization problem rather than directly predicting solutions. These consistent improvements indicate that the proposed framework generalizes effectively beyond the ER training distribution and remains robust to heterogeneous degree distributions, community structures, and small-world topology.

Due to space constraints, the ablation results for the Dual-GNN architecture are presented in Figure~\ref{fig:gnn_ablation} in the appendix. The state-space compression statistics across different graph
sizes are also reported in Table~\ref{tab:state_retention_sizes} in the
appendix.

\section{Conclusion}
\label{sec:conclusion}

We presented Dual-GNN Multilevel Coarsening, a learning-guided framework
for efficiently finding large independent sets. Rather than predicting a
solution directly on the original graph, our approach learns how to reduce
the optimization problem while preserving decisions that are useful for
the final MIS objective. The Partition GNN constructs a task-aligned
coarsening hierarchy, and the Representative GNN selects a compact set of
local independent-set states for coarse optimization. Completion-aware
search and deterministic recovery then translate the reduced solution back
to the original graph, guaranteeing that every returned solution is a valid
maximal independent set.

Future work could extend the multilevel coarsening framework to a broader range of combinatorial optimization problems. In particular, it would be valuable to investigate whether task-aligned coarsening strategies can be adapted to problems with different feasibility structures and objective functions, such as vertex cover, graph matching, and routing problems. Exploring such extensions could help assess the generality of the framework and further clarify the role of learned coarsening in reducing large combinatorial search spaces.

\clearpage
\subsection*{AI use statement}

In this work, we used generative AI tools to assist with language
editing, including improving grammar, clarity and readability.
These tools were not used to generate the research ideas, design the methodology,
conduct the experiments, analyze the results, or select and verify the cited
literature.
 All AI-assisted text was carefully reviewed and revised by the
authors to ensure its accuracy and consistency with the underlying research.
The authors take full responsibility for the final content of this work,
including all text, claims, results, and artifacts.

\bibliography{iclr2027_conference}
\bibliographystyle{iclr2027_conference}

\clearpage

\appendix
\section{Appendix}

\begin{figure}[H]
    \centering
    \includegraphics[width=1.0\linewidth]{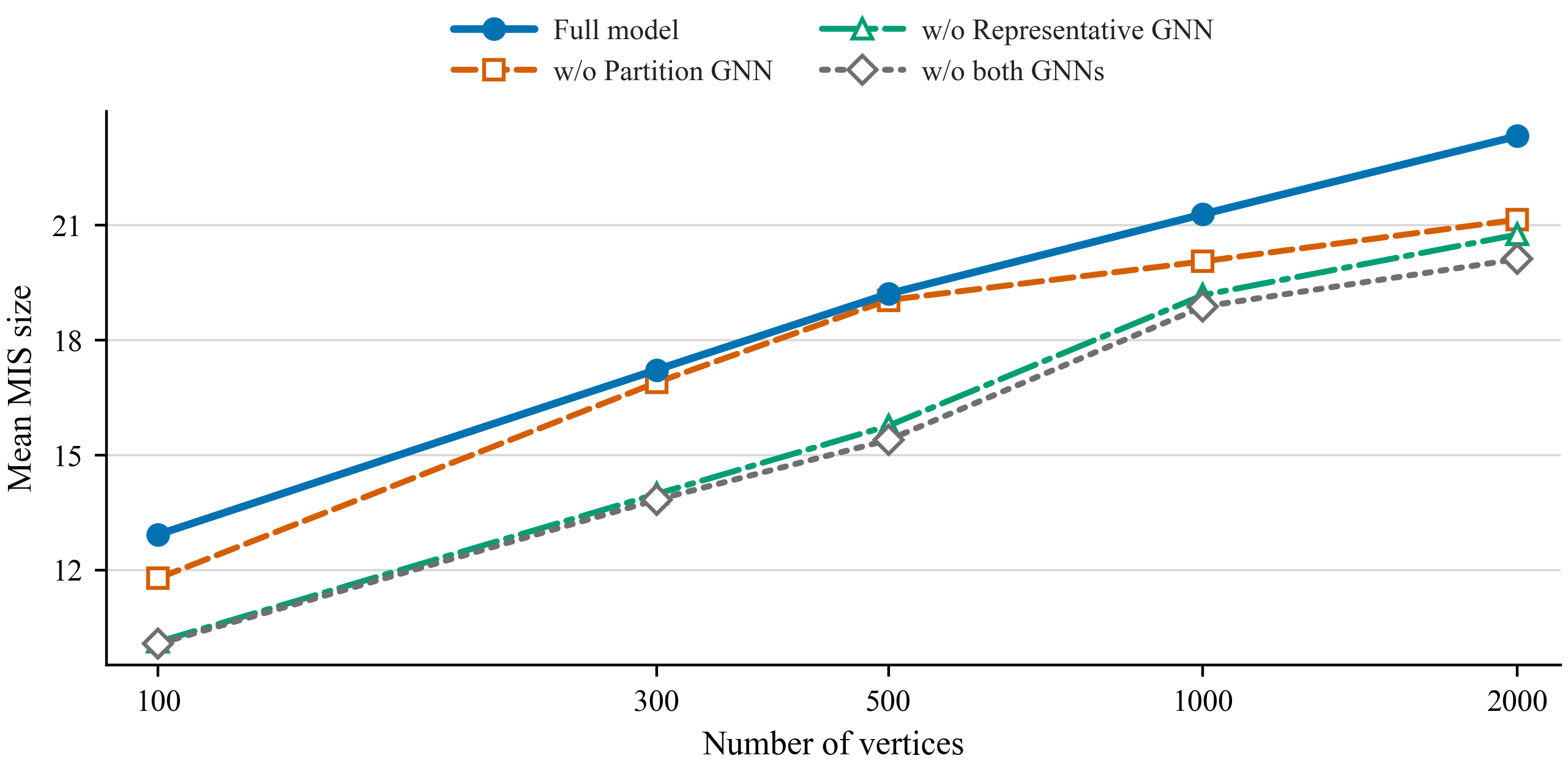}
    \caption{Ablation of the two GNN components}
    \label{fig:gnn_ablation}
\end{figure}

\paragraph{Ablation of the Dual-GNN Components.}
To examine the contribution of the two learned components, we compare the
full model with three ablated variants: removing the Partition GNN,
removing the Representative GNN, and removing both GNNs. When the
Partition GNN is removed, cluster contractions are determined using the
corresponding structural heuristic without learned guidance. When the
Representative GNN is removed, candidate local states are selected using
the non-learned scoring rule. All variants use the same test instances and
downstream optimization protocol as the full model. Figure~\ref{fig:gnn_ablation}
reports the mean MIS size on ER graphs with $p=0.35$ across different
graph sizes.

The full model consistently achieves the best performance across all graph
sizes, demonstrating the effectiveness of both GNN components. Removing the
Representative GNN leads to a larger performance drop, indicating the
importance of learned state selection. Removing the Partition GNN has a
smaller effect on medium-sized graphs, but the performance gap increases on 
larger graphs. Removing both GNNs produces the weakest overall results.
These results confirm that both GNN components play important roles in the proposed framework and are necessary for maintaining strong solution quality across different graph scales.

\paragraph{Runtime across Graph Densities.} Table~\ref{tab:er_density_runtime} reports the mean solving time of different methods on ER graphs under varying edge probabilities.

\begin{table}[H]
\centering
\caption{Mean solving time on ER graphs under different edge probabilities.
}
\label{tab:er_density_runtime}
\resizebox{\linewidth}{!}{%
\begin{tabular}{rcccccc}
\toprule
$p$
& Greedy
& Erd\H{o}s GNN
& Dynamic MIS
& pCQO-MIS
& ReduMIS
& Ours \\
\midrule
0.05
& $0.04\pm0.01$
& $0.14\pm0.06$
& $3.65\pm1.66$
& $4.97\pm1.26^{\dagger}$
& $5.68\pm1.18$
& $4.98\pm0.20$ \\

0.10
& $0.04\pm0.01$
& $0.18\pm0.05$
& $5.74\pm1.12$
& $5.20\pm1.33^{\dagger}$
& $5.56\pm0.17$
& $3.82\pm0.12$ \\

0.20
& $0.04\pm0.01$
& $0.52\pm0.12$
& $0.77\pm0.13$
& $5.51\pm1.47$
& $6.62\pm0.32$
& $3.26\pm0.35$ \\

0.35
& $0.07\pm0.01$
& $0.47\pm0.07$
& $1.42\pm0.61$
& $3.72\pm0.58$
& $12.91\pm1.23$
& $3.41\pm0.78$ \\

0.50
& $0.03\pm0.01$
& $0.94\pm0.19$
& $11.56\pm1.12$
& $5.78\pm0.92$
& $10.64\pm1.25$
& $5.10\pm0.99$ \\

0.70
& $0.04\pm0.01$
& $1.12\pm0.27$
& $10.94\pm1.76$
& $5.46\pm1.22$
& $15.72\pm2.18$
& $5.23\pm1.07$ \\
\bottomrule
\end{tabular}%
}
\vspace{2pt}
\parbox{\linewidth}{\footnotesize
$^{\dagger}$ pCQO-MIS returned empty, non-maximal solutions at
$p=0.05$ and $p=0.10$. The reported runtimes correspond to these
unsuccessful runs and therefore do not represent the time required
to obtain valid MIS solutions.
}
\end{table}
Our method maintains a relatively stable runtime across different graph
densities, ranging from 3.26 to 5.23 seconds. It is consistently faster
than ReduMIS, although lightweight greedy and some direct neural
baselines require less computation.

\paragraph{State-Space Compression across Graph Sizes.}
Table~\ref{tab:state_retention_sizes} reports the reduction from the
original graph to the final state-conflict graph for different graph
sizes.

\begin{table}[H]
\centering
\caption{State-space compression on ER graphs of different sizes.
}
\label{tab:state_retention_sizes}
\setlength{\tabcolsep}{8pt}
\renewcommand{\arraystretch}{1.12}

\begin{tabular}{rcccc}
\hline
$n$
& Final clusters 
& Candidate states 
& Retained states 
& Retention ratio \\
\hline
100  & 25.00  & 338.24   & 75.00  & 22.20\% \\
300  & 38.74  & 5457.20  & 116.22 & 2.44\% \\
500  & 63.00  & 10439.04 & 189.00 & 1.81\% \\
1000 & 127.50 & 23393.18 & 382.50 & 1.95\% \\
2000 & 250.00 & 35841.02 & 750.00 & 2.09\% \\
\hline
\end{tabular}
\end{table}

Although the number of candidate local states increases rapidly with
the graph size, the bounded state-selection mechanism keeps the final
conflict graph compact. For graphs with at least 300 vertices, only
approximately 2\% of the candidate states are retained for the final
optimization.

\paragraph{Sensitivity to the Representative-State Budget.} We select the representative-state budget $k$ using the held-out
validation set of 100 ER graphs with $n=500$ and $p=0.35$. We evaluate
$k\in\{1,2,3,5\}$ while fixing the trained models and all other inference
and search parameters. The adaptive state-allocation branch is disabled
in this experiment, so each final cluster retains at most the
$k$ highest-scoring states.
\begin{figure}[H]
    \centering
    \includegraphics[width=0.8\linewidth]{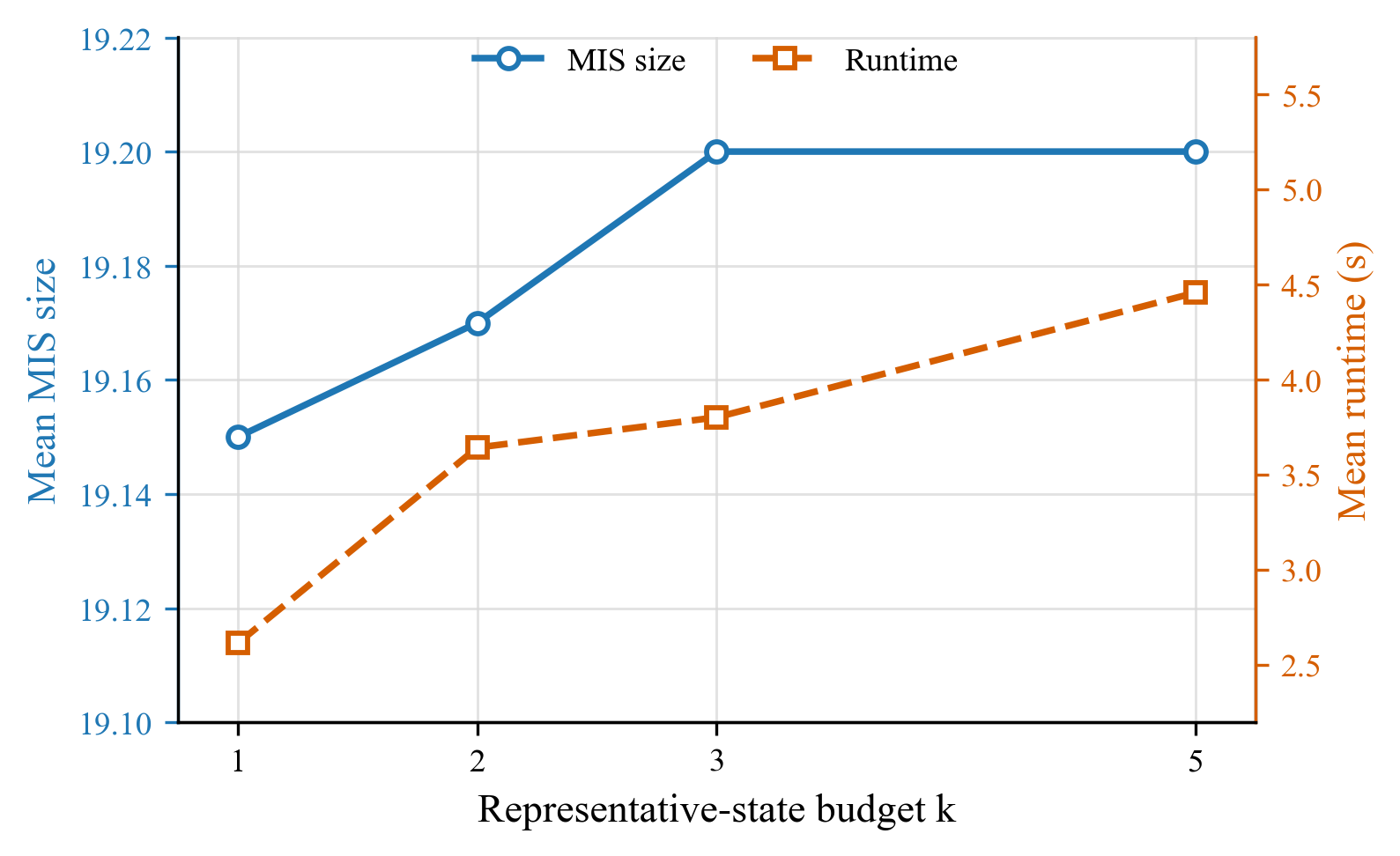}
    \caption{Sensitivity to the representative-state budget $k$}
    \label{fig:placeholder}
\end{figure}

As shown in Figure~\ref{fig:placeholder}, increasing the budget from
$k=1$ to $k=3$ improves the mean MIS size from 19.15 to 19.20, indicating
that retaining multiple local configurations benefits global coordination
during coarse optimization. Increasing the budget further to $k=5$ does
not improve the solution quality, while increasing the mean runtime from
3.80 to 4.46 seconds. We therefore use $k=3$ in all main experiments, as
it provides the best validation quality while avoiding the additional
computational cost associated with a larger retained state set.

\end{document}